%% file: main.tex
\documentclass[journal]{IEEEtran}

\usepackage{cite}
\usepackage{amsmath,amssymb,amsfonts}
\usepackage{graphicx}
\usepackage{booktabs}
\usepackage{multirow}
\usepackage{threeparttable}
\usepackage{array}
\usepackage{placeins}
\usepackage{flushend}
\usepackage{xcolor}
\usepackage{url}
\usepackage[hidelinks]{hyperref}
\hypersetup{
  pdftitle={When Semantics Saturate or Emerge: Adaptation-Conditional Semantic Utility in Source-Free Cross-Domain Few-Shot Learning},
  pdfauthor={Wei Liu, Xing Deng, Haijian Shao},
  pdfkeywords={source-free cross-domain few-shot learning, vision-language models, semantic prompting, low-rank adaptation}
}


\newcommand{\deltazero}{\Delta_{0}}
\newcommand{\deltalora}{\Delta_{L}}
\newcommand{\deltashift}{\Delta_{\mathrm{shift}}}

\title{When Semantics Saturate or Emerge: Adaptation-Conditional Semantic Utility in Source-Free Cross-Domain Few-Shot Learning}

\author{Wei Liu, XING DENG, HAIJIAN SHAO
\thanks{(Corresponding author: XING DENG.)}
\thanks{Wei Liu, XING DENG and HAIJIAN SHAO are with the College of Computer Science, Jiangsu University of Science and Technology, Zhenjiang 212003, China(e-mail: wliu@stu.just.edu.cn, xdeng@just.edu.cn, jsj\_shj@just.edu.cn, feiw@just.edu.cn )}}

\begin{document}
\maketitle

\begin{abstract}
Language descriptions in source-free cross-domain few-shot learning (SF-CDFSL) are often selected according to zero-shot accuracy obtained with a frozen vision--language model. This paper asks whether that ranking remains valid after target-domain visual adaptation. Under a strictly paired protocol, we compare a generic class-name template with fixed detailed class descriptions before and after visual Low-Rank Adaptation (LoRA) on EuroSAT, CropDisease, ISIC, and ChestX. Let $\deltazero$ and $\deltalora$ denote the Detailed-minus-Base accuracy before and after adaptation, respectively. Two recurring regimes emerge. In \emph{semantic saturation}, $\deltazero>0$ but $0<\deltalora\ll\deltazero$: on EuroSAT and CropDisease, initial gains of 8.13--21.54 percentage points contract to 0.69--2.96 points after LoRA. In \emph{semantic emergence}, $\deltazero\leq0$ but $\deltalora>0$: on ISIC and ChestX, detailed descriptions become more useful only after the visual representation is updated. Training trajectories and sample-level decomposition show that saturation is driven mainly by Base-LoRA recovering errors already solved by detailed semantics, whereas emergence is associated with prediction turnover and newly formed Detailed-only correct decisions. Fixed-point-free shuffled-semantic controls, a second CLIP backbone, and multiple random seeds support the broad pattern while identifying ChestX 1-shot as a weak boundary case. These findings establish that zero-shot prompt quality is an incomplete proxy for adaptation-anchor quality and motivate evaluating language on both sides of the adaptation boundary.
\end{abstract}


\IEEEpeerreviewmaketitle

\input{sections/01_introduction}
\input{sections/02_related_work}
\input{sections/03_problem_formulation}
\input{sections/04_experimental_setup}
\input{sections/05_main_observation}
\input{sections/06_training_dynamics}
\input{sections/07_sample_mechanisms}
\input{sections/08_controls_generalization}
\input{sections/09_discussion}
\input{sections/10_conclusion}

\FloatBarrier
\bibliographystyle{IEEEtran}
\bibliography{references}

\clearpage
\appendices
\input{sections/appendix_additional}
\FloatBarrier

\end{document}

%% file: sections/01_introduction.tex
\section{Introduction}
\IEEEPARstart{C}{ross-domain} few-shot learning (CDFSL) aims to recognize previously unseen classes in a target domain from only a small labeled support set. The challenge is especially pronounced when the target domain departs substantially from conventional natural-image pretraining data, as in the BSCD-FSL benchmark, which spans satellite, plant-disease, dermatological, and radiological imagery~\cite{guo2020broader}. Source-free CDFSL (SF-CDFSL) further removes access to the original source data, leaving the learner with only a pretrained model and the target-domain support examples~\cite{yazdanpanah2022visual,xu2024imdcl}. Under this restriction, successful adaptation must extract as much transferable structure as possible from the pretrained representation while avoiding overfitting to a handful of target samples.

Vision--language models (VLMs), particularly CLIP~\cite{radford2021clip}, provide a natural foundation for this setting because target classes can be represented by language and the visual encoder can be adapted with parameter-efficient mechanisms. Prompt learning~\cite{zhou2022coop,khattak2023maple}, adapters, and Low-Rank Adaptation (LoRA)~\cite{hu2022lora,zanella2024clip_lora} have therefore become common tools for few-shot VLM transfer. Within SF-CDFSL, recent studies have examined semantic prompting~\cite{zhuo2024prompt}, style-aware prompt tuning~\cite{xu2025stepspt}, attention degradation~\cite{yi2026attention,zhao2026reviving}, discriminability imbalance~\cite{zhang2026trap}, underused textual representations~\cite{zhang2026lost}, and local cross-modal alignment~\cite{zhao2026local}. These efforts share a broad objective: to exploit language as a semantic prior when the target-domain visual evidence is scarce.

A largely implicit assumption in this line of research is that a text description that improves frozen-model classification should also constitute a better target for few-shot adaptation. This assumption is intuitive but need not hold. Frozen zero-shot classification evaluates the compatibility between a \emph{fixed} visual representation and a text embedding. By contrast, visual LoRA changes the representation itself under supervision from the target support set. Consequently, language can interact with adaptation in at least two qualitatively different ways. A detailed description may provide a strong initial decision boundary whose benefit becomes redundant once the visual encoder learns from labeled target examples. Conversely, a description that is poorly aligned with the frozen visual space may still encode a useful class geometry that becomes accessible only after adaptation reshapes the target-domain representation. Zero-shot prompt quality and adaptation-anchor quality are therefore related, but not equivalent, quantities.

This paper isolates that distinction through a controlled comparison in which the adaptation algorithm, support set, optimization protocol, and visual backbone are fixed, while only the textual representation of the episode classes is changed. The \emph{Base} view uses a conventional class-name template, whereas the \emph{Detailed} view uses fixed class-specific visual descriptions. For each view, we evaluate both frozen CLIP and a separately trained visual LoRA. We define the zero-shot semantic utility
\begin{equation}
    \deltazero = A_D^0-A_B^0,
\end{equation}
and the adapted semantic utility
\begin{equation}
    \deltalora = A_D^L-A_B^L,
\end{equation}
where $A_B^0$ and $A_D^0$ denote frozen-model accuracies under the Base and Detailed views, and $A_B^L$ and $A_D^L$ denote the corresponding accuracies after visual LoRA. Their relationship reveals whether language acts primarily as an immediately usable prior or as an adaptation-dependent anchor.

Across EuroSAT, CropDisease, ISIC, and ChestX, two reproducible regimes emerge. In \emph{semantic saturation}, $\deltazero>0$ but $0<\deltalora\ll\deltazero$. EuroSAT and CropDisease obtain substantial frozen-model gains from the Detailed view, ranging from $8.13$ to $21.54$ percentage points, yet retain only $0.69$--$2.96$ points after LoRA. In \emph{semantic emergence}, $\deltazero\leq 0$ while $\deltalora>0$. On ISIC and ChestX, the Detailed view is not a better zero-shot classifier, but it becomes a better adaptation anchor after the visual representation is updated. These results show that the usefulness of language is \emph{adaptation-conditional}: the prompt ranking observed before adaptation does not reliably predict the ranking after adaptation.

We further examine why the two regimes arise. Training trajectories show that the large semantic advantage on EuroSAT and CropDisease contracts rapidly as Base-LoRA learns from the support set, whereas the advantage on ISIC and ChestX appears only after several adaptation steps. At the sample level, saturation is characterized by high retention of Detailed zero-shot correct samples together with high coverage of Detailed-only corrections by Base-LoRA, indicating substantial overlap between the errors solved by detailed semantics and those solved by visual adaptation. Emergence instead involves a broader reconstruction of the decision boundary rather than preservation of the frozen Detailed predictions. A fixed-point-free shuffled-semantic control confirms that these effects depend on the correct class--description correspondence rather than merely on longer text or an arbitrary text codebook. The same regime split persists with a second CLIP backbone and across multiple random seeds.

Accordingly, this work is positioned as an empirical and diagnostic study rather than a new adaptation algorithm. Its purpose is to establish what language quality means when the visual representation used for classification is itself trainable, and to provide the controls needed for future method papers to evaluate that question.

The contributions of this work are summarized as follows:
\begin{itemize}
    \item We introduce an adaptation-conditional evaluation framework that separates the semantic utility of a text view before and after visual adaptation, avoiding the common practice of treating frozen zero-shot accuracy as a proxy for adaptation usefulness.
    \item We identify two stable regimes in SF-CDFSL: \emph{semantic saturation}, in which a strong initial language advantage is largely absorbed by support-driven visual adaptation, and \emph{semantic emergence}, in which useful language structure becomes effective only after the visual representation adapts.
    \item We provide a sample-level and temporal account of these regimes through training trajectories, correction-overlap analysis, retention and coverage statistics, and prompt-transfer measurements.
    \item We establish the robustness of the finding through a shuffled-semantic control, two CLIP visual backbones, multiple random seeds, four heterogeneous target domains, and both 1-shot and 5-shot evaluation protocols.
\end{itemize}

%% file: sections/02_related_work.tex
\section{Related Work}

\subsection{Cross-Domain and Source-Free Few-Shot Learning}
Foundational few-shot learning develops transferable metrics, prototypes, or rapid adaptation rules from episodic training~\cite{vinyals2016matching,finn2017maml,snell2017proto,sung2018relation,chen2019closer,triantafillou2020metadataset}. Cross-domain few-shot learning (CDFSL) adds a mismatch between representation learning and target episodes. The BSCD-FSL study established the four-domain benchmark used here and showed that domain distance can dominate conventional few-shot difficulty~\cite{guo2020broader}. Later work studies feature-wise transformation, domain similarity, target-conditioned adapters, metric calibration, prototype revision, and source-side style diversification~\cite{tseng2020featurewise,oh2022understanding,li2022taskspecific,li2022ranking,zhou2023revisitingproto,fu2022wavesan,fu2023styleadv,zou2024flor}. These methods demonstrate that both source representation quality and target support structure determine transfer.

Source-free CDFSL removes access to source images during target adaptation. Visual Domain Bridge corrects internal distribution mismatch without revisiting source data~\cite{yazdanpanah2022visual}; IM-DCL combines information maximization with distance-aware contrastive learning on the target support set~\cite{xu2024imdcl}; and StepSPT uses step-wise distribution-aligned style prompts~\cite{xu2025stepspt}. Recent strong CDFSL work further investigates reconstruction targets, attention temperature, CLS-token behavior, image-token continuity, and random registers~\cite{ma2025damim,zou2024attntemp,zou2024cdcls,yi2025recit,yi2025reap}. This literature primarily asks how to improve adaptation. We instead hold adaptation fixed and ask how the value assigned to language changes across the adaptation boundary.

\subsection{Parameter-Efficient Adaptation of Vision--Language Models}
CLIP performs recognition by comparing image features with text-derived class prototypes~\cite{radford2021clip}. Its downstream behavior is therefore jointly determined by the visual representation and the linguistic form of each class. CoOp learns continuous textual context~\cite{zhou2022coop}, CoCoOp conditions prompts on inputs~\cite{zhou2022cocoop}, and MaPLe couples visual and textual prompts~\cite{khattak2023maple}. Knowledge-guided and regularized prompt learners preserve pretrained knowledge or improve unseen-class transfer~\cite{yao2023kgcoop,zhu2023prograd,khattak2023promptsrc}. Tip-Adapter, CLIP-Adapter, and TaskRes instead adapt a cache, residual feature interface, or classifier residual~\cite{zhang2022tipadapter,gao2024clipadapter,yu2023taskres}.

Low-Rank Adaptation injects trainable low-rank updates into frozen weight matrices~\cite{hu2022lora}. CLIP-LoRA established this strategy as a strong few-shot VLM baseline~\cite{zanella2024clip_lora}. Language itself can also be enriched with LLM-generated class descriptions, as in CuPL and description-based classifiers~\cite{pratt2023cupl,menon2023dclip}. These studies show that both text formulation and parameter-efficient adaptation matter, but typically compare prompts at one fixed model state. Our paired quantities $\deltazero$ and $\deltalora$ test whether a prompt ranking measured before visual adaptation remains valid after the representation moves.

\subsection{Language and Cross-Modal Mechanisms in SF-CDFSL}
Language-assisted SF-CDFSL has progressed from semantic-guided prompt diversity and style alignment~\cite{zhuo2024prompt,xu2025stepspt} to increasingly fine-grained analyses of CLIP adaptation. Recent work attributes limitations to exacerbated attention sink, a discriminability trap, underused intermediate text layers, degraded local alignment, or uniform treatment of semantically weak visual tokens~\cite{yi2026attention,zhang2026trap,zhang2026lost,zhao2026local,yi2026atha}. Semantic Probe contrasts prompt- and adapter-based tuning and introduces text-guided attention rectification~\cite{zhao2026reviving}.

Our study is complementary but differently positioned. It neither proposes another loss nor claims a unique internal failure mode. Visual LoRA is used as a controlled probe: the same target episode is optimized against two fixed text coordinate systems, and semantic value is measured before, during, and after that movement. This isolates whether language acts as an immediately useful prior, an adaptation-dependent anchor, or a prior whose relative advantage becomes redundant under support supervision.

%% file: sections/03_problem_formulation.tex
\section{Problem Formulation}
\label{sec:problem}

\subsection{Source-Free Episodic Recognition}
Let $f_{\theta}$ and $g_{\phi}$ denote the pretrained CLIP visual and text encoders, respectively. Both encoders are initialized from the same pretrained model, and no source-domain image or label is available during target adaptation. An $N$-way $K$-shot target episode is
\begin{equation}
    \mathcal{E}=(\mathcal{S},\mathcal{Q}),
\end{equation}
with labeled support set
\begin{equation}
    \mathcal{S}=\{(x_i^{s},y_i^{s})\}_{i=1}^{NK}
\end{equation}
and query set
\begin{equation}
    \mathcal{Q}=\{(x_j^{q},y_j^{q})\}_{j=1}^{NQ}.
\end{equation}
The query labels are used only for evaluation. All adaptation is performed independently within each episode using $\mathcal{S}$.

Each episode class $c\in\{1,\ldots,N\}$ is represented by two fixed textual views. The \emph{Base} view $b_c$ is a conventional class-name template, whereas the \emph{Detailed} view $d_c$ is a class-specific visual description. Their normalized text embeddings are
\begin{equation}
    t_c^{V}
    =\frac{g_{\phi}(p_c^{V})}
    {\lVert g_{\phi}(p_c^{V})\rVert_2},
    \qquad V\in\{B,D\},
\end{equation}
where $p_c^{B}=b_c$ and $p_c^{D}=d_c$. We collect the class anchors of view $V$ as
\begin{equation}
    \mathcal{T}_{V}=\{t_1^{V},\ldots,t_N^{V}\}.
\end{equation}
The two views describe the same class identities; only their linguistic specificity differs.

For visual parameters $\vartheta$, let
\begin{equation}
    z_{\vartheta}(x)
    =\frac{f_{\vartheta}(x)}{\lVert f_{\vartheta}(x)\rVert_2}
\end{equation}
be the normalized image representation. Classification with text view $V$ uses
\begin{equation}
    s_c^{V}(x;\vartheta)
    =\tau\, z_{\vartheta}(x)^{\top}t_c^{V},
\end{equation}
where $\tau$ is the CLIP logit scale. The frozen-model query accuracy associated with $\mathcal{T}_{V}$, averaged over the same paired episode distribution, is denoted by $A_V^{0}$.

\subsection{Text-View-Conditioned Visual Adaptation}
We adapt only the visual encoder with Low-Rank Adaptation (LoRA), while keeping the original CLIP weights and the entire text encoder frozen. For an adapted projection matrix $W_{\ell}$ at layer $\ell$,
\begin{equation}
    W_{\ell}'=W_{\ell}+\Delta W_{\ell},
    \qquad
    \Delta W_{\ell}=\mathbf{B}_{\ell}\mathbf{A}_{\ell},
\end{equation}
where $\mathbf{A}_{\ell}$ and $\mathbf{B}_{\ell}$ are trainable low-rank factors~\cite{hu2022lora}. Let $\Delta\theta$ collect all trainable LoRA parameters.

The same support episode induces two controlled adaptation runs, one for each text view. For training view $U\in\{B,D\}$, the LoRA parameters are obtained by
\begin{equation}
\begin{aligned}
    \Delta\theta_{U}^{\star}
    =\arg\min_{\Delta\theta}\frac{1}{|\mathcal{S}|}
    \sum_{(x_i^{s},y_i^{s})\in\mathcal{S}}
    \operatorname{CE}\big(&\mathbf{s}^{U}(x_i^{s};
    \theta\oplus\Delta\theta),\\[-1mm]
    &y_i^{s}\big),
\end{aligned}
\label{eq:view_conditioned_lora}
\end{equation}
where $\theta\oplus\Delta\theta$ denotes the frozen CLIP visual parameters augmented by the LoRA update, and $\mathbf{s}^{U}$ is the vector of logits defined by $\mathcal{T}_{U}$. The Base and Detailed runs use the same classes, support/query samples, initialization protocol, data stream, and optimization schedule; only the text anchor set used by the support loss changes.

To make the conditioning explicit, we denote by
\begin{equation}
    A_{U\rightarrow V}^{L}
\end{equation}
the mean query accuracy obtained after training LoRA with text view $U$ and evaluating the adapted visual encoder with text view $V$. The matched-view accuracies used in the principal comparison are
\begin{equation}
    A_B^{L}=A_{B\rightarrow B}^{L},
    \qquad
    A_D^{L}=A_{D\rightarrow D}^{L}.
\end{equation}
The cross-view quantities $A_{B\rightarrow D}^{L}$ and $A_{D\rightarrow B}^{L}$ measure prompt-coordinate dependence and are analyzed separately. This notation distinguishes the text view that \emph{induces} the visual update from the text view that \emph{reads out} the adapted representation.

\subsection{Adaptation-Conditional Semantic Utility}
We quantify the relative value of the Detailed view against the Base view at two stages of the pipeline. Before adaptation, the zero-shot semantic utility is
\begin{equation}
    \deltazero=A_D^{0}-A_B^{0}.
\end{equation}
After view-matched visual adaptation, the adapted semantic utility is
\begin{equation}
    \deltalora=A_D^{L}-A_B^{L}.
\end{equation}
Their change across the adaptation boundary is
\begin{equation}
    \deltashift=\deltalora-\deltazero.
\end{equation}
All three quantities are reported in percentage points. They are comparative and protocol dependent: they measure the utility of one text view relative to another for a fixed backbone, target domain, shot, and adaptation procedure, rather than assigning an intrinsic quality score to a prompt.

The experiments reveal two recurring regimes. We define \emph{semantic saturation} by
\begin{equation}
    \deltazero>0,
    \qquad
    0<\deltalora<\deltazero,
\label{eq:saturation}
\end{equation}
which implies $\deltashift<0$. In this regime, the Detailed view is already beneficial in the frozen representation, but support-driven visual adaptation makes part of that advantage redundant. For conditions satisfying~\eqref{eq:saturation}, the absorbed fraction can be summarized as
\begin{equation}
    \rho_{\mathrm{sat}}
    =1-\frac{\deltalora}{\deltazero}.
\end{equation}

We define \emph{semantic emergence} by
\begin{equation}
    \deltazero\leq 0,
    \qquad
    \deltalora>0,
\label{eq:emergence}
\end{equation}
so that $\deltashift>0$. Here, the Detailed view is not superior under the frozen visual representation, yet becomes the better anchor after the support set reshapes that representation. Its emergence magnitude is naturally measured by $\deltashift$.

These two regimes are the empirically observed cases central to this work; they are not intended as an exhaustive taxonomy of every possible sign pattern. Statistical confidence in each assignment is assessed with paired episode-level bootstrap intervals, described in Sec.~IV-C.

%% file: sections/04_experimental_setup.tex
\section{Experimental Setup}
\label{sec:setup}

\subsection{Benchmarks and Episodic Protocol}
We evaluate on the four target domains introduced by BSCD-FSL~\cite{guo2020broader}: EuroSAT, CropDisease, ISIC, and ChestX. These benchmarks cover satellite scene recognition, plant disease classification, dermoscopic lesion recognition, and chest radiography, respectively, and therefore provide substantially different visual statistics and degrees of compatibility with the pretrained CLIP representation.

All experiments follow a source-free 5-way episodic protocol. Each episode contains either one or five labeled support images per class and 15 query images per class. Query images and labels are never used for adaptation. The principal results use 800 independently sampled episodes for 1-shot evaluation and 400 episodes for 5-shot evaluation. Within every episode, the Base and Detailed conditions share the same class subset, support samples, query samples, LoRA initialization, support training tensors, mini-batch order, and random-number stream. Consequently, each reported Detailed-minus-Base difference is an episode-wise paired comparison rather than a difference between independently sampled runs.

The two shot settings use an identical model and adaptation rule; only the number of labeled support images changes. Unless otherwise stated, the main tables report the standard seed-1 protocol above. All accuracies are averaged over episodes and expressed in percentage points.

\subsection{Controlled Adaptation Configuration}
\subsubsection{Text views}
The Base view uses the conventional template ``\texttt{a photo of a \{\}.}'' with the corresponding class name. The Detailed view uses a fixed bank of class-specific visual descriptions. The description bank is held constant throughout the study: no text prompt is generated, selected, ensembled, or optimized within an episode. Thus, the experiment evaluates the behavior of two already available text views rather than a prompt-generation procedure.

For a sampled 5-way episode, only the five embeddings associated with its selected classes are used. The CLIP text encoder is frozen in every condition. Base and Detailed runs differ solely in the text anchors supplied to the support classification loss and, for matched-view evaluation, to the final classifier. Cross-view evaluation reuses the same adapted visual model with the alternative text anchor set, yielding the complete prompt-transfer matrix defined in Sec.~\ref{sec:problem}.

\subsubsection{Visual backbone and LoRA}
The principal backbone is CLIP ViT-B/16~\cite{radford2021clip}. We insert LoRA modules into the query, key, and value projections of every visual Transformer block. The rank is 16, the LoRA scaling factor is 1, and the LoRA dropout rate is 0.25. All pretrained CLIP parameters, including the complete text encoder, remain frozen; only the inserted visual LoRA factors are trainable. A second set of experiments repeats the principal comparison with CLIP ViT-B/32 while leaving all remaining settings unchanged.

Each episode is optimized independently with AdamW using a learning rate of $10^{-4}$, weight decay $10^{-2}$, and cosine annealing to a minimum learning rate of $10^{-6}$. The support mini-batch size is at most 25, the evaluation batch size is 128, the input resolution is $224\times224$, and the cosine-similarity logits use a fixed scale of 100. Mixed-precision training is enabled on CUDA. EuroSAT, CropDisease, and ISIC use 250 optimizer updates per episode. Following the established baseline implementation, ChestX uses 125 updates. This dataset-specific budget is fixed before analysis and is identical for Base and Detailed conditions, preserving the paired comparison within every ChestX episode. No hyperparameter is tuned separately for either text view or either shot setting.

\subsection{Evaluation Suites and Statistical Analysis}
\subsubsection{Principal comparison}
For every episode, we first evaluate the frozen visual encoder with both text views, obtaining $A_B^{0}$ and $A_D^{0}$. We then train two visual-LoRA models from the same initialization: one with Base anchors and one with Detailed anchors. Each adapted model is evaluated with both anchor sets, producing $A_{B\rightarrow B}^{L}$, $A_{B\rightarrow D}^{L}$, $A_{D\rightarrow B}^{L}$, and $A_{D\rightarrow D}^{L}$. The matched-view quantities define the zero-shot and adapted semantic utilities, while the cross-view quantities quantify prompt-coordinate dependence. We additionally store per-query predictions and margins for the sample-level retention, forgetting, and correction analyses.

\subsubsection{Training dynamics and semantic controls}
Training dynamics are measured on 100 episodes for each dataset and shot. For the 250-update datasets, query and support statistics are recorded at updates $0,1,5,10,25,50,100,150,$ and $250$; for ChestX, the final checkpoint is update 125. At each checkpoint, both the Base-trained and Detailed-trained visual models are read out with both text views. This protocol measures when semantic utility changes without altering the optimization trajectory.

To verify that the observed effects depend on correct class semantics rather than merely on description length or on the geometry of an alternative text codebook, we construct a shuffled-semantic control. Within each episode, a fixed-point-free permutation reassigns the five Detailed embeddings to incorrect classes. The permutation preserves the set of text vectors while destroying class--description correspondence. Correct-Detailed and shuffled-Detailed conditions use the same episode and paired training stream. This control is evaluated on 100 episodes for every dataset and shot.

Generalization is assessed in two complementary ways. First, the complete Base/Detailed comparison is repeated with ViT-B/32 on 100 episodes per dataset and shot. Second, seeds 2 and 3 repeat all four datasets with 200 episodes for 1-shot and 100 episodes for 5-shot; these runs are summarized together with the principal seed-1 result. The additional runs test whether the sign and regime of semantic utility are stable across both model capacity and episodic randomness, rather than serving as a new hyperparameter search.

\subsubsection{Uncertainty estimation}
All inferential comparisons are paired at the episode level. We use 10,000 paired bootstrap resamples to estimate 95\% percentile confidence intervals for $\deltazero$, $\deltalora$, $\deltashift$, and the sample-level conditional rates. Two-sided paired bootstrap $p$-values are computed from the mass of the resampled difference distribution on either side of zero. We bootstrap episodes rather than pooling all query images, thereby respecting the dependence among samples drawn from the same episode. Regime assignments are based jointly on the observed signs and their paired confidence intervals; the corresponding operational definitions are given in Sec.~\ref{sec:problem}.

%% file: sections/05_main_observation.tex
\section{Main Observation: Semantic Utility Saturates or Emerges}
\input{tables/table1_main_semantic_utility.tex}

Table~\ref{tab:main_semantic_utility} reports the matched-view results under the controlled protocol of Section~\ref{sec:setup}. Across all eight dataset--shot conditions, Detailed-LoRA attains higher query accuracy than Base-LoRA. However, the same ordering does not hold before adaptation. The relation between zero-shot utility $\deltazero$ and adapted utility $\deltalora$ therefore separates the target domains into two qualitatively different regimes. Full paired-bootstrap confidence intervals are reported in Table~\ref{tab:main_semantic_utility_ci} of the appendix.

\begin{figure*}[t]
    \centering
    \includegraphics[width=0.92\textwidth]{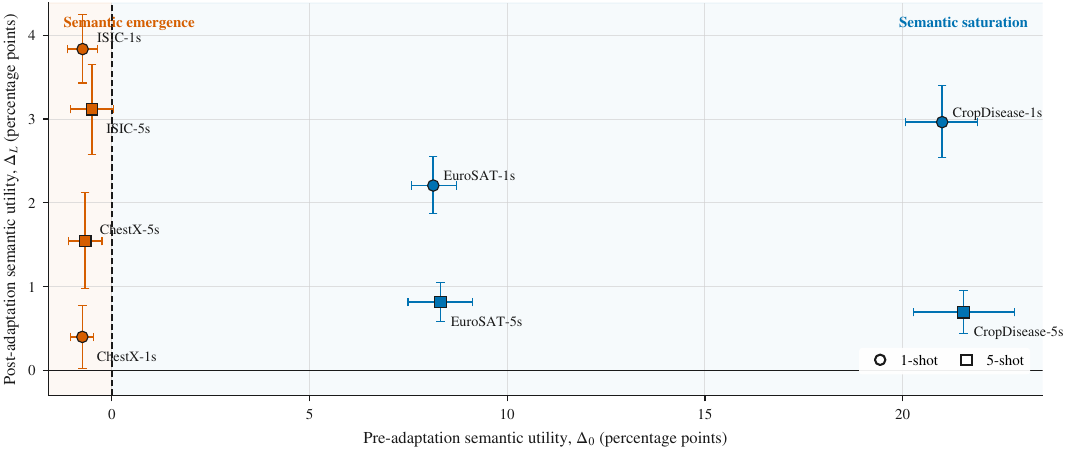}
    \caption{Adaptation-conditional semantic utility. The horizontal axis measures Detailed-minus-Base utility before visual LoRA ($\deltazero$), and the vertical axis measures the corresponding matched-view utility after LoRA ($\deltalora$). Circles and squares denote 1-shot and 5-shot results, respectively. EuroSAT and CropDisease fall in the saturation regime, whereas ISIC and ChestX fall in the emergence regime. Error bars show 95\% confidence intervals from 10,000 paired episode-level bootstrap resamples.}
    \label{fig:regime_map}
\end{figure*}

\subsection{Semantic Saturation}
EuroSAT and CropDisease exhibit a strong zero-shot benefit from Detailed descriptions. On EuroSAT, $\deltazero$ is $+8.13$ points in 1-shot episodes and $+8.31$ points in 5-shot episodes. CropDisease shows an even larger advantage of $+21.00$ and $+21.54$ points. These gains are statistically reliable, with 95\% confidence intervals that remain well above zero.

After visual LoRA, the Detailed view remains superior, but its relative advantage contracts sharply. EuroSAT retains only $+2.21$ points in 1-shot and $+0.82$ points in 5-shot, while CropDisease retains $+2.96$ and $+0.69$ points. Equivalently, adaptation removes approximately $72.9\%$ and $90.2\%$ of the initial EuroSAT advantage, and $85.9\%$ and $96.8\%$ of the initial CropDisease advantage. The negative shifts are significant in all four conditions.

This contraction should not be interpreted as a failure of Detailed-LoRA. Detailed-LoRA remains the best matched-view model in every saturation condition. The result is instead relative: supervised target-domain adaptation substantially improves the Base condition and makes much of the initial language advantage no longer exclusive to the Detailed view. The stronger contraction in 5-shot episodes further suggests that additional labeled support reduces dependence on zero-shot text quality, although the remaining positive $\deltalora$ shows that the two text views are not fully interchangeable.

\subsection{Semantic Emergence}
ISIC and ChestX show the opposite pattern. In frozen CLIP, the Detailed view provides no positive advantage. For ISIC, $\deltazero$ is $-0.74$ points in 1-shot and $-0.51$ points in 5-shot; the latter difference is statistically indistinguishable from zero because its confidence interval slightly crosses zero. ChestX exhibits negative zero-shot utilities of $-0.75$ and $-0.67$ points.

Visual adaptation reverses this ordering. Detailed-LoRA exceeds Base-LoRA by $+3.84$ and $+3.12$ points on ISIC, and by $+0.40$ and $+1.54$ points on ChestX. The resulting utility shifts are all positive: $+4.58$ and $+3.62$ points for ISIC, and $+1.15$ and $+2.22$ points for ChestX. The post-adaptation ChestX gain is comparatively small in 1-shot episodes, but remains positive under paired bootstrap testing ($p=0.034$).

Semantic emergence therefore cannot be explained by preserving an already superior frozen-model decision rule. The Detailed view becomes useful only after the support set has altered the visual representation. In these domains, the quality of a text view as an adaptation anchor is not revealed by its zero-shot accuracy.

\subsection{A Boundary Between Prompt Evaluation and Adaptation}
Figure~\ref{fig:regime_map} makes the distinction explicit. Saturation occupies the region $\deltazero>0$ and $0<\deltalora<\deltazero$, whereas emergence occupies the region $\deltazero\leq0$ and $\deltalora>0$. The two regimes are not separated by whether Detailed descriptions ultimately help: they improve matched-view LoRA accuracy in both. They are separated by \emph{when} that utility becomes observable.

This finding establishes the central empirical claim of the paper: frozen-model prompt quality is not a reliable proxy for adaptation-anchor quality. A text view may offer a large advantage that becomes largely redundant after few-shot adaptation, or it may appear unhelpful before adaptation and become beneficial only after the visual space is updated. The following sections examine how these two regimes develop during optimization and which sample-level transitions account for them.

%% file: tables/table1_main_semantic_utility.tex
\begin{table*}[t]
\centering
\begin{threeparttable}
\caption{Adaptation-conditional semantic utility on ViT-B/16 (seed 1). Detailed descriptions exhibit either saturation or emergence after visual LoRA.}
\label{tab:main_semantic_utility}
\footnotesize
\setlength{\tabcolsep}{3.0pt}
\renewcommand{\arraystretch}{1.10}
\begin{tabular}{llrrrrrrrl}
\toprule
\multirow{2}{*}{Dataset} & \multirow{2}{*}{Shot} & \multicolumn{2}{c}{Zero-shot accuracy} & \multicolumn{2}{c}{After visual LoRA} & \multicolumn{3}{c}{Semantic utility (pp)} & \multirow{2}{*}{Regime} \\
\cmidrule(lr){3-4}\cmidrule(lr){5-6}\cmidrule(lr){7-9}
 & & Base & Detailed & Base & Detailed & $\Delta_0$ & $\Delta_L$ & $\Delta_{\mathrm{shift}}$ & \\
\midrule
\multirow{2}{*}{EuroSAT} & 1 & 54.85 & \textbf{62.98} & 82.19 & \textbf{84.40} & +8.13 & +2.21 & -5.92 & \textsc{Sat.} \\
 & 5 & 54.55 & \textbf{62.86} & 92.74 & \textbf{93.55} & +8.31 & +0.82 & -7.49 & \textsc{Sat.} \\
\addlinespace[2pt]
\multirow{2}{*}{CropDisease} & 1 & 40.90 & \textbf{61.91} & 82.48 & \textbf{85.44} & +21.00 & +2.96 & -18.04 & \textsc{Sat.} \\
 & 5 & 40.80 & \textbf{62.34} & 94.80 & \textbf{95.49} & +21.54 & +0.69 & -20.85 & \textsc{Sat.} \\
\addlinespace[2pt]
\multirow{2}{*}{ISIC} & 1 & \textbf{27.24} & 26.50 & 35.43 & \textbf{39.27} & -0.74 & +3.84 & +4.58 & \textsc{Emer.} \\
 & 5 & \textbf{27.08} & 26.57 & 50.55 & \textbf{53.67} & -0.51 & +3.12 & +3.62 & \textsc{Emer.} \\
\addlinespace[2pt]
\multirow{2}{*}{ChestX} & 1 & \textbf{21.93} & 21.18 & 21.50 & \textbf{21.90} & -0.75 & +0.40 & +1.15 & \textsc{Emer.} \\
 & 5 & \textbf{21.84} & 21.17 & 23.06 & \textbf{24.60} & -0.67 & +1.54 & +2.22 & \textsc{Emer.} \\
\bottomrule
\end{tabular}
\begin{tablenotes}[flushleft]
\footnotesize
\item Accuracies and semantic-utility terms are percentage points. Bold denotes the better text view within the same adaptation state.
\item $\Delta_0=A_D^0-A_B^0$, $\Delta_L=A_D^L-A_B^L$, and $\Delta_{\mathrm{shift}}=\Delta_L-\Delta_0$.
\item \textsc{Sat.}: semantic saturation; \textsc{Emer.}: semantic emergence.
\end{tablenotes}
\end{threeparttable}
\end{table*}

%% file: sections/06_training_dynamics.tex
\section{How Semantic Utility Evolves During Adaptation}
\label{sec:dynamics}
\begin{figure*}[t]
    \centering
    \includegraphics[width=\textwidth]{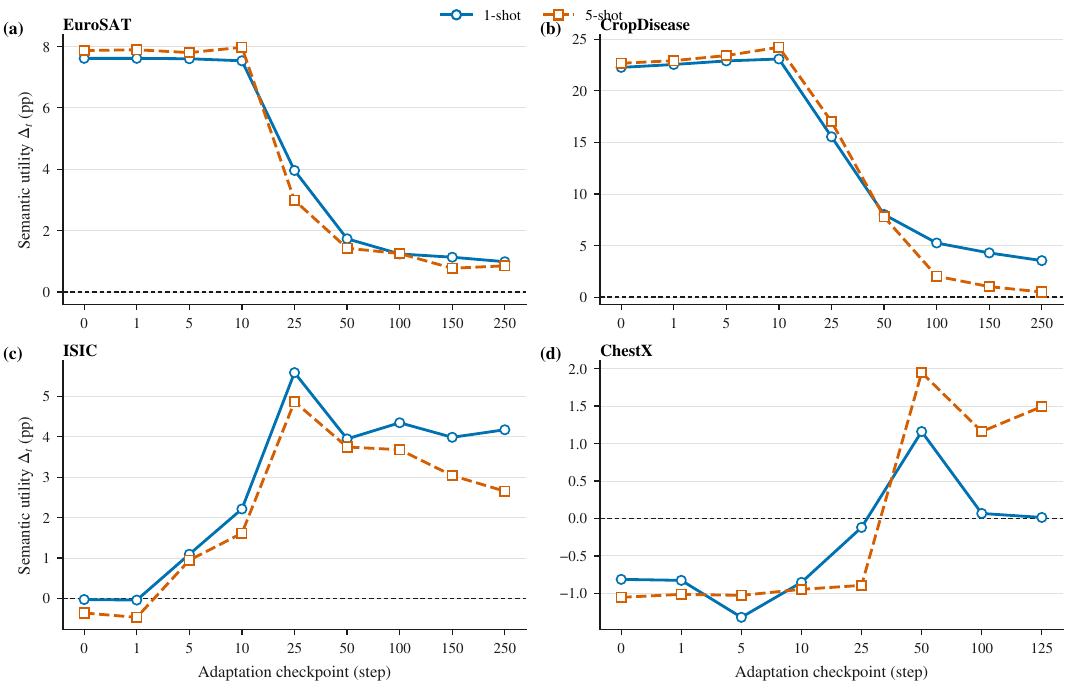}
    \caption{Evolution of matched-view semantic utility during visual LoRA adaptation. At checkpoint $t$, $\Delta_t=A_D(t)-A_B(t)$ compares Detailed-LoRA and Base-LoRA evaluated with their respective training text views. Step 0 corresponds to frozen CLIP. Curves report means over 100 paired episodes; consequently, their endpoints need not exactly equal the larger-scale estimates in Table~\ref{tab:main_semantic_utility}. EuroSAT and CropDisease show delayed but pronounced saturation, whereas ISIC and ChestX reveal early and late semantic emergence, respectively.}
    \label{fig:dynamics}
\end{figure*}

The final results establish that semantic utility can either saturate or emerge, but they do not reveal \emph{when} the two regimes form. We therefore evaluate the same paired episodes at intermediate adaptation checkpoints. Let
\begin{equation}
A_B(t)=A_{B\rightarrow B}^{L}(t),
\qquad
A_D(t)=A_{D\rightarrow D}^{L}(t),
\end{equation}
and define the checkpoint-dependent semantic utility as
\begin{equation}
\Delta_t=A_D(t)-A_B(t).
\label{eq:dynamic_utility}
\end{equation}
At $t=0$, no LoRA update has been applied and $\Delta_t$ reduces to the zero-shot utility on the dynamic-evaluation episodes. Figure~\ref{fig:dynamics} shows that the final regimes arise from distinct optimization trajectories rather than from a uniform decay or preservation of the initial prompt gap.

\subsection{Saturation Is Delayed, Not Immediate}
EuroSAT preserves almost all of its initial Detailed advantage during the first ten updates. In 1-shot episodes, $\Delta_t$ remains near $7.6$ points from step 0 through step 10; the 5-shot trajectory is similarly stable at approximately $7.9$ points. The sharp contraction occurs only after the model begins to fit the support set more substantially: at steps 25 and 50, the 1-shot utility falls to $3.96$ and $1.73$ points, while the 5-shot utility falls to $2.99$ and $1.43$ points. By the final checkpoint, only $0.99$ and $0.85$ points remain.

CropDisease makes the delayed nature of saturation even clearer. Detailed utility first \emph{increases}: from $22.27$ to $23.08$ points in 1-shot episodes and from $22.67$ to $24.21$ points in 5-shot episodes by step 10. It then contracts rapidly, reaching $8.01$ and $7.79$ points at step 50 and ending at $3.55$ and $0.51$ points. Thus, adaptation does not immediately suppress a useful semantic prior. During the earliest updates, the Detailed anchor can benefit at least as much as the Base anchor; saturation begins when support supervision becomes strong enough for Base-LoRA to recover class distinctions that were initially available only through richer language.

The matched-view accuracies support this differential-catch-up interpretation. Between steps 10 and 50, Base-LoRA gains $22.05$--$34.91$ points across the four EuroSAT/CropDisease conditions, whereas Detailed-LoRA gains $13.76$--$21.87$ points. The relative utility contracts because the Base condition improves faster, not because Detailed-LoRA collapses. This temporal evidence is consistent with the large-scale endpoint result: few-shot supervision progressively substitutes for part of the discriminative information supplied by Detailed text.

\subsection{Emergence Requires Movement of the Visual Space}
ISIC follows the opposite trajectory. Its Detailed utility is non-positive at initialization, but the ordering reverses after only five updates: $\Delta_5=+1.09$ points in 1-shot episodes and $+0.95$ points in 5-shot episodes. Utility then grows to $+5.59$ and $+4.87$ points at step 25 before settling at $+4.17$ and $+2.65$ points at convergence. Over the first 25 updates, Detailed-LoRA gains $8.15$ and $10.25$ accuracy points, compared with only $2.53$ and $5.03$ points for Base-LoRA. The Detailed text is therefore not merely revealed by longer training; it guides a more favorable early reorganization of the target visual representation.

ChestX exhibits a weaker and later transition. Both shots remain negative through step 25, and the sign reversal appears around step 50. The 5-shot trajectory remains positive thereafter and ends at $+1.49$ points. The 1-shot trajectory briefly reaches $+1.16$ points at step 50 but returns close to zero in the 100-episode dynamic subset. This fluctuation is consistent with the much smaller $+0.40$-point gain measured by the larger 800-episode main experiment: ChestX 1-shot is a weak-emergence case whose estimated magnitude is sensitive to episodic sampling, rather than a stable multi-point separation like ISIC. The delayed crossing also agrees with the severe visual mismatch between web-pretrained CLIP features and radiographic images.

\subsection{Regimes Reflect Relative Learning Rates}
The trajectories reveal three properties that are hidden by endpoint evaluation. First, semantic utility is non-monotonic: CropDisease is initially amplified before saturating, while ISIC peaks early and later settles to a smaller positive advantage. Second, the sign of zero-shot utility does not constrain the sign after adaptation; a text view that is initially neutral or unfavorable can become the better anchor once the visual encoder moves. Third, shot count changes the speed and stability of this transition rather than inducing one universal effect. More support accelerates saturation on EuroSAT and CropDisease and produces a clearer late emergence on ChestX, but it does not uniformly increase the peak utility on ISIC.

These observations refine the interpretation of adaptation-conditional utility. The relevant quantity is not how much semantic information a prompt contains in isolation, but how the two text-conditioned optimization processes improve relative to one another. Saturation occurs when Base-LoRA learns faster and closes an initial gap; emergence occurs when Detailed-LoRA extracts greater benefit from the same support supervision and opens a new gap. The next section examines the sample-level prediction transitions underlying these two temporal patterns.

%% file: sections/07_sample_mechanisms.tex
\section{Sample-Level Mechanisms}
\label{sec:sample_mechanisms}
The aggregate utilities in Sections~\ref{sec:dynamics} and the preceding endpoint analysis reveal whether Detailed text is relatively more useful, but not \emph{which query decisions carry that utility}. We therefore trace the correctness state of every query sample under Base-ZS, Detailed-ZS, Base-LoRA, and Detailed-LoRA. This analysis separates two possibilities that are indistinguishable from accuracy alone: adaptation may preserve an initially useful Detailed decision rule while Base-LoRA catches up, or it may replace the frozen decisions and create a new Detailed advantage.

For a text view $V\in\{B,D\}$ and adaptation state $s\in\{0,L\}$, let
\begin{equation}
\mathcal{C}_{V}^{s}
=
\{x\in\mathcal{Q}:\hat{y}_{V}^{s}(x)=y(x)\}
\end{equation}
denote the set of query samples classified correctly. We first measure how much of the Detailed zero-shot evidence survives adaptation:
\begin{equation}
R_D
=
\frac{|\mathcal{C}_{D}^{0}\cap\mathcal{C}_{D}^{L}|}
{|\mathcal{C}_{D}^{0}|}.
\label{eq:detailed_retention}
\end{equation}
Because a low retention rate can coexist with substantial correction of previously wrong samples, we also define
\begin{equation}
K_D
=
\frac{|(\mathcal{Q}\setminus\mathcal{C}_{D}^{0})\cap\mathcal{C}_{D}^{L}|}
{|\mathcal{Q}\setminus\mathcal{C}_{D}^{0}|},
\label{eq:detailed_correction}
\end{equation}
which is the Detailed-ZS error-correction rate of Detailed-LoRA.

To quantify how much of the initial Detailed advantage becomes reproducible from support labels alone, define the Detailed-only zero-shot set
\begin{equation}
\mathcal{U}_{D}^{0}
=
\mathcal{C}_{D}^{0}\setminus\mathcal{C}_{B}^{0},
\end{equation}
and its Base-LoRA coverage
\begin{equation}
O_B
=
\frac{|\mathcal{U}_{D}^{0}\cap\mathcal{C}_{B}^{L}|}
{|\mathcal{U}_{D}^{0}|}.
\label{eq:base_coverage}
\end{equation}
Finally, the adapted utility admits an exact sample-space decomposition. Let
\begin{equation}
X_D=\frac{|\mathcal{C}_{D}^{L}\setminus\mathcal{C}_{B}^{L}|}{|\mathcal{Q}|},
\qquad
X_B=\frac{|\mathcal{C}_{B}^{L}\setminus\mathcal{C}_{D}^{L}|}{|\mathcal{Q}|}.
\end{equation}
Then
\begin{equation}
\deltalora=X_D-X_B,
\label{eq:exclusive_decomposition}
\end{equation}
when accuracies are expressed as proportions. Thus, Detailed-LoRA is better precisely when its exclusive correct set exceeds that of Base-LoRA.

\begin{figure*}[t]
    \centering
    \includegraphics[width=0.92\textwidth]{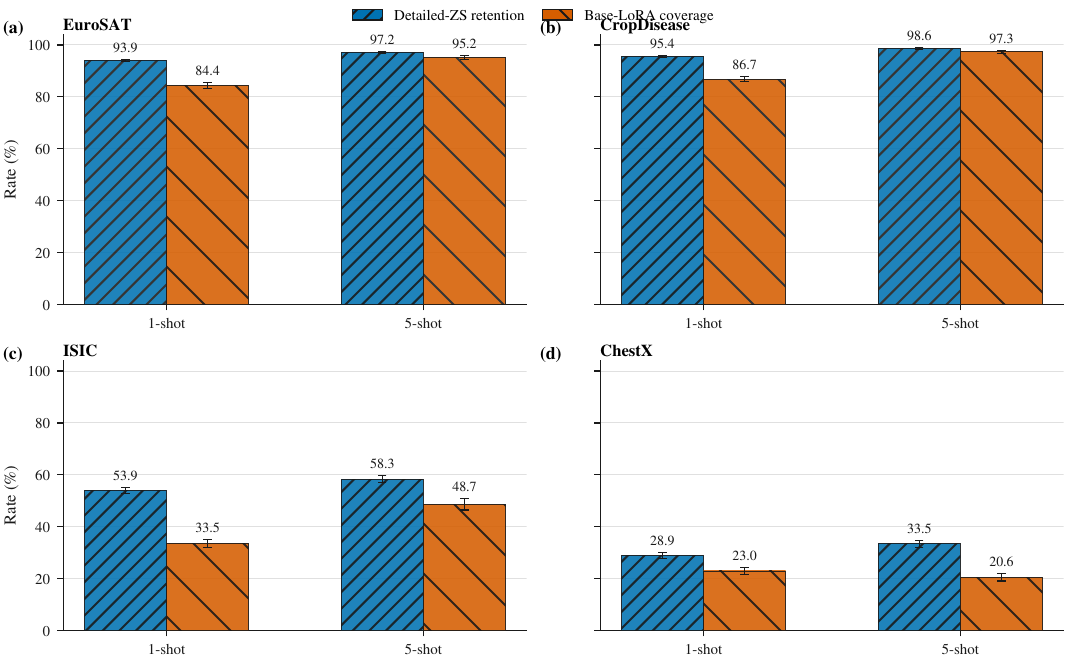}
    \caption{Sample-level evidence for the two semantic-utility regimes. Detailed-ZS retention measures the fraction of frozen Detailed successes preserved by Detailed-LoRA, while Base-LoRA coverage measures how often support supervision recovers samples that were correct only under Detailed-ZS. Error bars are 95\% confidence intervals from paired episode-level bootstrap resampling. Saturation domains exhibit high retention and high coverage; emergence domains exhibit substantially lower values, indicating greater decision turnover and less recovery of the initial Detailed-only evidence.}
    \label{fig:sample_mechanisms}
\end{figure*}

\subsection{Saturation Reflects Preservation Plus Overlap}
EuroSAT and CropDisease show that semantic saturation is not caused by Detailed-LoRA discarding its initial advantage. Detailed-ZS retention ranges from $93.9\%$ to $98.6\%$, while Detailed-LoRA also corrects $68.2\%$--$90.3\%$ of the samples that Detailed-ZS initially misclassifies. The Detailed condition therefore both preserves most of its frozen successes and expands beyond them after adaptation.

The contraction of relative utility is instead explained by cross-view overlap. Base-LoRA covers $84.4\%$--$97.3\%$ of the Detailed-only zero-shot set, with the largest coverage in 5-shot episodes. In other words, most samples whose correctness initially depends on richer language cease to be exclusive evidence once the Base condition can learn from labeled support. This sample-level recovery directly matches the temporal catch-up observed in Section~\ref{sec:dynamics}.

The exclusive-correct decomposition makes the remaining advantage explicit. On EuroSAT, Detailed-LoRA is exclusively correct on $5.8\%$ and $2.4\%$ of query samples, compared with $3.6\%$ and $1.6\%$ for Base-LoRA, yielding the final $+2.21$- and $+0.82$-point utilities. CropDisease similarly yields $7.3\%-4.3\%=+2.96$ points in 1-shot and $2.3\%-1.6\%=+0.69$ points in 5-shot. Most post-adaptation successes are therefore shared, and the initial semantic advantage saturates because Base-LoRA reproduces much of the same discriminative outcome.

\subsection{Emergence Reflects Decision Turnover}
ISIC and ChestX exhibit a different transition. Detailed-ZS retention is only $53.9\%$--$58.3\%$ on ISIC and $28.9\%$--$33.5\%$ on ChestX. These values should not be interpreted as harmful forgetting: the frozen Detailed classifier is not superior in these domains. Instead, adaptation replaces many initial decisions while correcting new samples. Detailed-LoRA corrects $34.0\%$ and $52.0\%$ of Detailed-ZS errors on ISIC, and $20.0\%$ and $22.2\%$ on ChestX. The resulting predictor is therefore not a preserved version of Detailed-ZS, but a substantially revised decision rule supported by the target examples.

Base-LoRA also covers much less of the Detailed-only zero-shot set in these domains: $33.5\%$--$48.7\%$ on ISIC and $20.6\%$--$23.0\%$ on ChestX. More importantly, the final Detailed advantage is created by a new imbalance in exclusive correctness. On ISIC, Detailed-LoRA-only correctness reaches $11.4\%$ and $10.9\%$, versus $7.6\%$ and $7.8\%$ for Base-LoRA, exactly accounting for the $+3.84$- and $+3.12$-point utilities. ChestX shows a much narrower imbalance: $11.2\%$ versus $10.8\%$ in 1-shot and $13.8\%$ versus $12.2\%$ in 5-shot. This near balance explains why ChestX emergence is positive but weaker, particularly in the 1-shot setting.

\subsection{Inherited Versus Reconstructed Semantic Evidence}
The two regimes can now be distinguished at the sample level. In saturation domains, Detailed semantic evidence is largely \emph{inherited}: Detailed-LoRA retains the frozen successes, but Base-LoRA learns many of the same cases from support supervision, leaving only a small exclusive advantage. In emergence domains, useful Detailed evidence is largely \emph{reconstructed}: the frozen Detailed successes are unstable, yet adaptation creates more new Detailed-only correct samples than Base-only ones.

The complete transition statistics and the algebraic derivation of Eq.~\eqref{eq:exclusive_decomposition} are reported in Appendix~D.

This decomposition does not identify a unique internal feature, attention path, or gradient component responsible for each transition. It establishes a stricter empirical boundary: saturation is dominated by preservation and overlap, whereas emergence is dominated by prediction turnover and newly formed exclusive correctness. The next section tests whether these effects depend on genuine class semantics and whether they generalize across backbones and random seeds.

%% file: sections/08_controls_generalization.tex
\section{Controls and Generalization}
\label{sec:controls}
\begin{figure*}[t]
    \centering
    \includegraphics[width=0.90\textwidth]{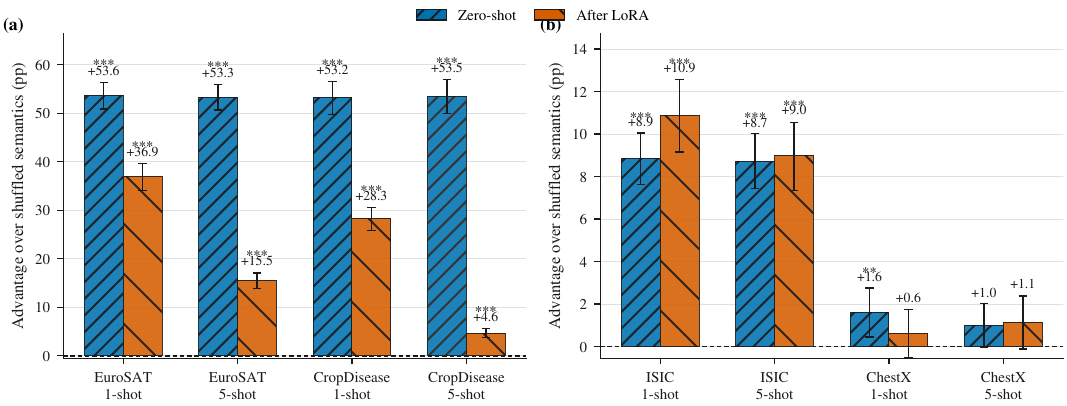}
    \caption{Aligned Detailed accuracy minus a fixed-point-free shuffled Detailed codebook. The control preserves the text-embedding multiset but breaks class--description correspondence. Error bars are 95\% paired-bootstrap intervals.}
    \label{fig:shuffle}
\end{figure*}

\textbf{Shuffled semantics.} Let $\pi(c)\neq c$ permute the episode classes and assign $\widetilde{\mathbf t}_c^D=\mathbf t_{\pi(c)}^D$. Correct alignment strongly outperforms this codebook on EuroSAT, CropDisease, and ISIC, before and after LoRA (Fig.~\ref{fig:shuffle}). The effect cannot therefore be reduced to prompt length or merely replacing the text-vector basis. Yet Shuffled-LoRA becomes surprisingly competent in 5-shot episodes, showing that labels can partially organize visual features around even an incorrect codebook. ChestX differences are small and not significant after LoRA, consistent with its weak semantic signal. Full intervals are in Appendix~E.

\textbf{Backbone and seed robustness.} ViT-B/32 preserves saturation on EuroSAT/CropDisease and adaptation-created utility on ISIC/ChestX. ISIC 1-shot changes from formal emergence to amplification because its ViT-B/32 $\deltazero$ is slightly positive, illustrating that the robust object is the direction of utility shift rather than an immutable categorical label. Across three seeds, seven of eight conditions retain the exact regime; only the near-zero ChestX 1-shot case changes label. Appendix~E reports all results.

\textbf{Prompt transfer.} Cross-view evaluation shows $A_{B\rightarrow B}^L\neq A_{B\rightarrow D}^L$ and $A_{D\rightarrow D}^L\neq A_{D\rightarrow B}^L$. Visual LoRA is therefore prompt-conditioned rather than a freely composable domain correction. Non-zero transfer gaps do not by themselves imply harmful overfitting; the matched Detailed view remains superior. Appendix~F reports the full matrix.

%% file: sections/09_discussion.tex
\section{Discussion}
\label{sec:discussion}
\textbf{What prompt quality measures.} $\deltazero$ measures compatibility with the frozen visual representation, whereas $\deltalora$ measures the usefulness of an anchor while that representation is optimized. Their mismatch means prompt quality is not an intrinsic scalar attached to a sentence. Zero-shot evaluation remains informative, but it is an incomplete proxy for adaptation utility.

\textbf{How labels interact with semantics.} In saturation domains, support supervision substitutes for part of the initially accessible semantic prior: Base-LoRA learns overlapping corrections. In emergence domains, labels first reorganize the visual space, after which richer anchors become useful. Language and supervision are therefore neither globally additive nor globally redundant. Methods that force preservation of frozen predictions may be harmful in emergence, while methods that assume semantic and adaptation gains add independently ignore the large overlap in saturation.

\textbf{Evaluation and positioning.} Language-assisted SF-CDFSL should report $(\deltazero,\deltalora,\deltashift)$ on paired episodes and include cross-prompt evaluation when the visual module is trained against text anchors. This paper is an empirical/diagnostic contribution, not a state-of-the-art method claim. Appendix~G places Detailed-LoRA beside representative published results. Its accuracy is competitive with vanilla CLIP-LoRA and several source-free baselines but below the strongest recent optimization methods on average. Direct ranking is secondary because backbone, episode count, augmentation, update budget, language resources, and target-image access differ.

\textbf{Scope.} The evidence is limited to fixed class-level descriptions, visual LoRA, two CLIP-B backbones, four targets, and 5-way 1/5-shot episodes. The analysis identifies prediction overlap and turnover but not a unique internal causal component. The shuffled control establishes the importance of correct class--description mapping, not which words or visual attributes are causal. Detailed descriptions constitute additional class-level target-domain side information constructed from an audited reference pool that is disjoint from the reported episodic images. Accordingly, the published-method comparison is contextual rather than a harmonized state-of-the-art test. Within the controlled Base--Detailed analysis, however, the episode data and optimization protocol are identical, isolating the effect of the fixed text view. Future work should predict utility regimes from short trajectories and test adaptation-aware prompt selection across additional adaptation families under harmonized information access.

%% file: sections/10_conclusion.tex
\section{Conclusion}
\label{sec:conclusion}
Detailed class semantics do not have a fixed value across frozen and adapted CLIP representations. Under paired visual-LoRA experiments, EuroSAT and CropDisease show semantic saturation: a large zero-shot advantage becomes largely shared with Base-LoRA. ISIC and, more weakly, ChestX show semantic emergence: detailed anchors become useful only after visual adaptation. Training trajectories, sample transitions, shuffled semantics, prompt transfer, a second backbone, and multiple seeds support this adaptation-conditional view. The practical conclusion is direct: prompt evaluation for SF-CDFSL should occur on both sides of the adaptation boundary, and visual LoRA should be interpreted as co-adapting with the text coordinates used for training.

%% file: sections/appendix_additional.tex
\section{Reproducibility and Information Access}
\label{app:reproducibility}

\subsection{Target Domains and Episode Construction}
The four target datasets follow the BSCD-FSL evaluation family: EuroSAT for remote-sensing scenes~\cite{helber2019eurosat}, CropDisease for plant pathology~\cite{mohanty2016plant}, ISIC for dermoscopic lesions~\cite{codella2019isic}, and ChestX for thoracic radiography~\cite{wang2017chestx}. Every episode samples five target classes, $K\in\{1,5\}$ labeled support images per class, and 15 disjoint query images per class. No source-domain image is loaded during episode adaptation. Base and Detailed conditions consume the same episode object and are paired before any model update.

The principal ViT-B/16 run contains 800 episodes for 1-shot and 400 for 5-shot. Dynamics and shuffled controls use 100 paired episodes per dataset and shot. ViT-B/32 uses 100 paired episodes, while the additional seed runs use 200 episodes for 1-shot and 100 for 5-shot. These smaller suites are diagnostic replications and are not substituted for the principal endpoint estimates.

\subsection{Detailed Description Bank and Information Access}
Each class is represented by one fixed, class-level visual description. The descriptions were constructed offline with Qwen3.5-27B-FP8 from one to three representative reference images per class. The generation instruction restricted the output to observable, class-defining visual evidence, including appearance, structure, color, texture, and spatial organization, while excluding image-specific incidental details and unsupported encyclopedic knowledge. After construction, the description bank was frozen and reused unchanged across all episodes. No episodic support or query image, model prediction, gradient, or evaluation outcome was used to generate, select, or revise a description.

The reference images were maintained in a separate reference pool and excluded from every reported support and query set. We verified this separation against the serialized episode manifests using canonical image identifiers and file-content hashes. Detailed therefore introduces additional \emph{class-level target-domain side information} relative to the class-name Base condition, but it does not use episode-specific or query-dependent information. The Base and Detailed conditions remain strictly paired: they use identical classes, support/query images, initialization, optimization, and random streams, with only the fixed text view changed. Comparisons with published methods are consequently reported as contextual rather than harmonized rankings whenever the permitted language or image side information differs across studies.

\subsection{Optimization and Numerical Configuration}
\begin{table}[t]
\centering
\caption{Principal visual-LoRA configuration.}
\label{tab:implementation_details}
\footnotesize
\setlength{\tabcolsep}{4pt}
\renewcommand{\arraystretch}{1.08}
\begin{tabular}{p{0.30\columnwidth}p{0.62\columnwidth}}
\toprule
Component & Setting \\
\midrule
Backbone & CLIP ViT-B/16; replication with ViT-B/32 \\
Trainable modules & Visual Q/K/V LoRA in all Transformer blocks \\
Rank / scale / dropout & 16 / 1 / 0.25 \\
Frozen modules & Original visual weights and complete text encoder \\
Optimizer & AdamW, learning rate $10^{-4}$, weight decay $10^{-2}$ \\
Schedule & Cosine decay to $10^{-6}$ \\
Updates & 250; ChestX 125 \\
Support batch & At most 25 \\
Evaluation batch & 128 \\
Input / logit scale & $224\times224$ / 100 \\
Precision & CUDA mixed precision \\
\bottomrule
\end{tabular}
\end{table}

All tensors participating in similarity or matrix multiplication are placed on the same concrete device and converted to compatible floating-point types. This is essential because pretrained CLIP modules may execute in FP16 while newly inserted LoRA factors default to FP32. Base and Detailed runs use identical data order, initialization, augmentation, optimizer state construction, and update count.

\subsection{Paired Inference and Bootstrap Procedure}
For each episode $e$, every reported contrast is first computed as a within-episode difference, e.g., $d_e=A_{D,e}^{L}-A_{B,e}^{L}$. We draw 10,000 bootstrap samples of episodes with replacement and recompute the mean difference. The 2.5th and 97.5th percentiles form the reported 95\% interval~\cite{efron1994bootstrap}. Two-sided $p$-values are obtained from twice the smaller bootstrap tail probability and clipped at one. Query images are not pooled as independent observations because samples within an episode share classes, support data, and an adapted model.

\section{Full Endpoint Statistics and Regime Taxonomy}
\label{app:endpoint}
\input{tables/table1_main_semantic_utility_with_ci.tex}

The sign-based labels in the main text cover the two recurring patterns, but the framework admits a broader taxonomy:
\begin{table}[t]
\centering
\caption{Complete qualitative taxonomy of adaptation-conditional utility.}
\label{tab:regime_taxonomy}
\footnotesize
\setlength{\tabcolsep}{3.5pt}
\renewcommand{\arraystretch}{1.08}
\begin{tabular}{p{0.36\columnwidth}p{0.20\columnwidth}p{0.30\columnwidth}}
\toprule
Condition & Name & Interpretation \\
\midrule
$\Delta_0>0$, $0<\Delta_L<\Delta_0$ & Saturation & Initial benefit contracts \\
$\Delta_0>0$, $\Delta_L>\Delta_0$ & Amplification & Initial benefit strengthens \\
$\Delta_0\leq0$, $\Delta_L>0$ & Emergence & Benefit appears after adaptation \\
$\Delta_0>0$, $\Delta_L\leq0$ & Reversal & Initial benefit is lost/reversed \\
$\Delta_0\leq0$, $\Delta_L\leq0$ & Persistent non-benefit & Detailed never becomes superior \\
\bottomrule
\end{tabular}
\end{table}

A discrete label should not substitute for uncertainty. ISIC 5-shot has a slightly negative point estimate before LoRA but a confidence interval that touches zero; ChestX 1-shot has a small positive post-LoRA estimate whose magnitude varies across seeds. We preserve their operational labels for consistency while explicitly treating near-zero conditions as boundaries.

The taxonomy is deliberately threshold-free: a regime name follows the signs and ordering of $(\Delta_0,\Delta_L)$, whereas evidential strength is conveyed separately by the paired interval, $p$-value, and effect magnitude. This prevents an arbitrary ``practical significance'' cutoff from changing the scientific definition. For comparison across domains, we report absolute percentage-point differences rather than relative percentage changes, because the latter would inflate effects on low-accuracy domains such as ChestX. The continuous quantities remain primary; the regime label is a compact summary of their relationship.

\FloatBarrier
\section{Additional Training-Dynamics Evidence}
\label{app:dynamics}
The combined curve in Fig.~\ref{fig:dynamics} is reproduced below as individual domain panels to expose scale differences that are compressed in the shared layout.

\begin{figure*}[t]
\centering
\begin{minipage}{0.49\textwidth}\centering
\includegraphics[width=\linewidth]{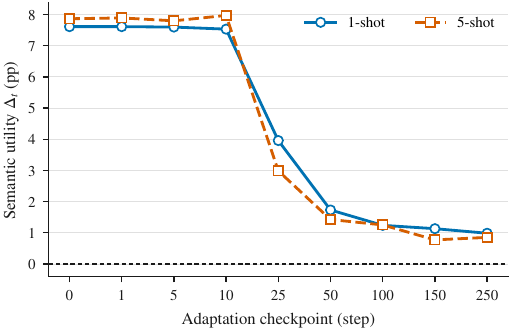}
\end{minipage}\hfill
\begin{minipage}{0.49\textwidth}\centering
\includegraphics[width=\linewidth]{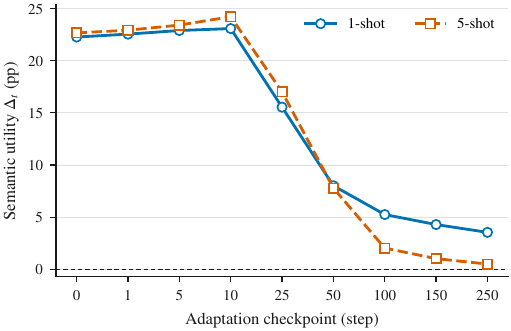}
\end{minipage}

\vspace{1mm}
\begin{minipage}{0.49\textwidth}\centering
\includegraphics[width=\linewidth]{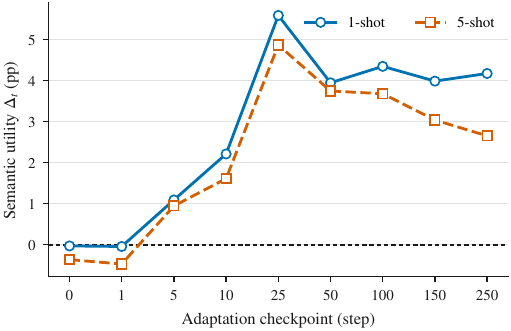}
\end{minipage}\hfill
\begin{minipage}{0.49\textwidth}\centering
\includegraphics[width=\linewidth]{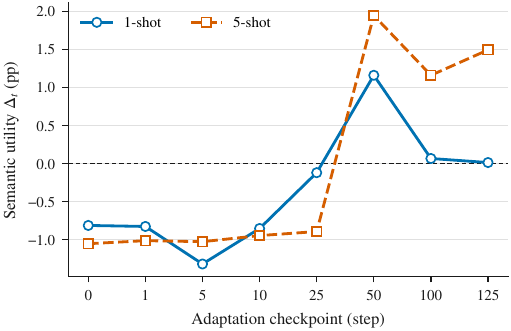}
\end{minipage}
\caption{Per-domain training trajectories. EuroSAT and CropDisease exhibit delayed contraction after an initially stable or amplified Detailed advantage. ISIC crosses into positive utility within the first few updates, whereas ChestX crosses later and remains weakest in 1-shot.}
\label{fig:dynamics_individual}
\end{figure*}

The dynamic suite uses 100 paired episodes and therefore estimates a trajectory rather than reproducing the exact principal endpoint. The decisive observation is directional: the large saturation-domain gap closes because Base-LoRA improves relatively faster, while the emergence-domain gap becomes positive only after the visual representation has moved. No checkpoint is selected using query accuracy; all checkpoints are predetermined and evaluated retrospectively.

Each trajectory can be read through four descriptive quantities: its initial utility $\Delta_0$, the first observed sign crossing, the maximum absolute utility reached during adaptation, and the final utility. These summaries are not used for model selection. In particular, the temporary peak on ISIC and the early amplification on CropDisease are retained rather than replaced by the best checkpoint. This distinction is necessary because the paper studies how utility evolves under a fixed adaptation schedule, not how to tune a stopping rule with query labels. The paired checkpoint curves also reduce a common ambiguity in training plots: a changing gap may arise from Base improving, Detailed improving, or both. The endpoint and sample-level analyses show that the dominant source differs across regimes.
\FloatBarrier

\section{Complete Sample-Level Decomposition}
\label{app:sample}
\input{tables/table2_sample_mechanisms.tex}

For completeness, Eq.~\eqref{eq:exclusive_decomposition} follows from partitioning the two correct sets into their intersection and exclusive parts:
\begin{align}
A_D^L-A_B^L
&=\frac{|\mathcal C_D^L|-|\mathcal C_B^L|}{|\mathcal Q|}\\
&=\frac{|\mathcal C_D^L\setminus\mathcal C_B^L|
-|\mathcal C_B^L\setminus\mathcal C_D^L|}{|\mathcal Q|}\\
&=X_D-X_B.
\end{align}
The shared intersection cancels exactly. Thus, the adapted utility is not merely correlated with exclusive correctness; it is identical to the difference between the two exclusive-correct rates.

The conditional rates in Table~\ref{tab:sample_mechanisms} answer distinct questions. Retention asks whether Detailed-LoRA preserves frozen Detailed successes. Error correction asks whether it expands beyond those successes. Base-LoRA coverage asks whether support supervision can recover the cases initially unique to richer language. Their joint pattern separates preservation-plus-overlap from decision reconstruction without asserting a unique internal attention or gradient mechanism.
\FloatBarrier

\section{Semantic Controls, Backbones, and Seeds}
\label{app:controls}
\input{tables/table4_shuffle_control.tex}
\input{tables/table5_backbone_generalization.tex}
\input{tables/table6_multiseed_summary.tex}

The shuffled condition is intentionally stronger than a random-text baseline: it preserves the exact Detailed embedding multiset and all pairwise distances up to relabeling. It therefore isolates class--description correspondence. Its limitation is equally important: a fixed-point-free permutation changes every class anchor simultaneously and does not reveal which phrase, visual attribute, or class pair drives the aligned advantage.

The control separates three claims that are often conflated. First, a richer text-embedding set is not sufficient: assigning the same vectors to the wrong classes sharply reduces performance on EuroSAT, CropDisease, and ISIC. Second, the non-zero shuffled-LoRA accuracy shows that support supervision can partially learn even an incorrect codebook, especially with five shots. Third, the aligned--shuffled difference after adaptation measures the value of correct correspondence under a fixed text multiset; it does not measure the total value of language relative to a text-free classifier. ChestX remains the important weak case, where aligned--shuffled post-LoRA intervals include zero.

Across ViT-B/16 and ViT-B/32, EuroSAT/CropDisease consistently contract while ISIC/ChestX gain relative utility after adaptation. The exact categorical label can change when $\Delta_0$ lies close to zero, as in ISIC 1-shot on ViT-B/32. The backbone study therefore supports a directional generalization claim---contraction versus adaptation-created utility---rather than invariance of every discrete label.

The three-seed aggregation addresses a different source of uncertainty. It does not enlarge the number of principal episodes or replace the paired bootstrap; instead, it tests whether the sign pattern survives independently generated episode streams. Seven of eight dataset--shot conditions retain the same regime across all seeds. ChestX 1-shot is the sole exception and has a mean $\Delta_L$ close to zero, so its instability is consistent with effect size rather than evidence against the broader phenomenon. These results motivate reporting continuous utilities and uncertainty in addition to regime names.
\FloatBarrier

\section{Prompt-Transfer Matrix}
\label{app:transfer}
\input{tables/table3_prompt_transfer.tex}

\begin{figure}[t]
\centering
\includegraphics[width=\columnwidth]{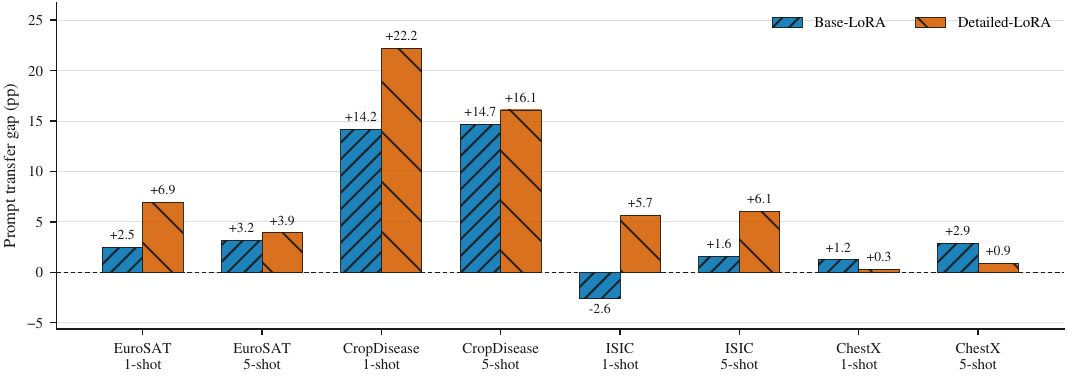}
\caption{Prompt-transfer gaps after Base-conditioned and Detailed-conditioned visual LoRA. Non-zero gaps show that the visual update is coupled to the anchor coordinates used during support optimization. A transfer gap is a diagnostic of co-adaptation, not by itself evidence of harmful overfitting.}
\label{fig:prompt_transfer}
\end{figure}

Define the signed transfer gaps as $G_B=A_{B\rightarrow B}^{L}-A_{B\rightarrow D}^{L}$ and $G_D=A_{D\rightarrow D}^{L}-A_{D\rightarrow B}^{L}$. Positive values mean that the prompt used during support optimization remains preferable at evaluation; a negative value means that switching prompts improves accuracy. The latter occurs for Base-trained ISIC 1-shot and prevents us from equating every non-zero gap with a loss.

CropDisease exhibits the largest coordinate dependence: changing the test prompt after training changes accuracy by 14--22 points. EuroSAT and ISIC have smaller but clear signed gaps, while ChestX is weakly prompt-conditioned. The asymmetry $G_B\neq G_D$ also matters: Base-conditioned and Detailed-conditioned LoRA do not learn a single prompt-neutral domain correction viewed from two test templates. Instead, each update is optimized against a different anchor geometry. This observation justifies matched-view reporting, but it does not imply that prompt dependence is harmful overfitting; that judgment still requires query accuracy and generalization evidence. Visual LoRA should therefore not be treated as a universal target-domain residual that can be trained under one text space and freely recombined with another.
\FloatBarrier

\section{Contextual Comparison with Published Accuracy Results}
\label{app:comparison}
\input{tables/table7_contextual_comparison.tex}

Tables~\ref{tab:contextual_comparison} and~\ref{tab:contextual_comparison_5shot} are included because an empirical analysis must demonstrate that its observations occur in a meaningful performance regime. Detailed-LoRA is competitive with the reported CLIP-LoRA baseline and reaches the best contextual EuroSAT result in both shots and the best 1-shot ISIC result among the listed rows. It is not the best average method: ATHA and Semantic Probe explicitly optimize final adaptation and are stronger overall, especially on 5-shot ISIC/ChestX or CropDisease. This is consistent with the paper's positioning rather than a deficiency in the central claim.

The comparison is not used as a formal superiority test. IM-DCL uses a different backbone, and the CLIP methods vary in episode count, optimization length, augmentation, prompt expansion, and text-side information. Recent Semantic Probe experiments, for example, evaluate 400 episodes in both shot settings and incorporate LLM-expanded text, whereas ATHA uses 800/400 episodes and a 100-epoch schedule. Our numbers come from the paired protocol described in Appendix~\ref{app:reproducibility}. A fair ranking would require all methods to be rerun with the same episodes, text bank, image-access policy, and optimization budget.

%% file: tables/table1_main_semantic_utility_with_ci.tex
\begin{table*}[t]
\centering
\begin{threeparttable}
\caption{Paired episode-level bootstrap analysis of semantic utility.}
\label{tab:main_semantic_utility_ci}
\footnotesize
\setlength{\tabcolsep}{3.0pt}
\renewcommand{\arraystretch}{1.08}
\begin{tabular}{llrrrrrr}
\toprule
\multirow{2}{*}{Dataset} & \multirow{2}{*}{Shot} & \multicolumn{2}{c}{Before LoRA} & \multicolumn{2}{c}{After LoRA} & \multicolumn{2}{c}{Adaptation shift} \\
\cmidrule(lr){3-4}\cmidrule(lr){5-6}\cmidrule(lr){7-8}
 & & $\Delta_0$ [95\% CI] & $p$ & $\Delta_L$ [95\% CI] & $p$ & $\Delta_{\mathrm{shift}}$ [95\% CI] & $p$ \\
\midrule
\multirow{2}{*}{EuroSAT} & 1 & +8.13 [+7.57, +8.71] & $<.0001$ & +2.21 [+1.87, +2.55] & $<.0001$ & -5.92 [-6.55, -5.29] & $<.0001$ \\
 & 5 & +8.31 [+7.49, +9.11] & $<.0001$ & +0.82 [+0.58, +1.05] & $<.0001$ & -7.49 [-8.33, -6.64] & $<.0001$ \\
\addlinespace[2pt]
\multirow{2}{*}{CropDisease} & 1 & +21.00 [+20.07, +21.90] & $<.0001$ & +2.96 [+2.54, +3.40] & $<.0001$ & -18.04 [-19.00, -17.09] & $<.0001$ \\
 & 5 & +21.54 [+20.28, +22.82] & $<.0001$ & +0.69 [+0.44, +0.96] & $<.0001$ & -20.85 [-22.14, -19.56] & $<.0001$ \\
\addlinespace[2pt]
\multirow{2}{*}{ISIC} & 1 & -0.74 [-1.12, -0.35] & 0.0002 & +3.84 [+3.43, +4.24] & $<.0001$ & +4.58 [+4.06, +5.08] & $<.0001$ \\
 & 5 & -0.51 [-1.05, +0.05] & 0.072 & +3.12 [+2.57, +3.65] & $<.0001$ & +3.62 [+2.83, +4.41] & $<.0001$ \\
\addlinespace[2pt]
\multirow{2}{*}{ChestX} & 1 & -0.75 [-1.04, -0.45] & $<.0001$ & +0.40 [+0.03, +0.78] & 0.034 & +1.15 [+0.68, +1.60] & $<.0001$ \\
 & 5 & -0.67 [-1.09, -0.25] & 0.002 & +1.54 [+0.98, +2.13] & $<.0001$ & +2.22 [+1.51, +2.91] & $<.0001$ \\
\bottomrule
\end{tabular}
\begin{tablenotes}[flushleft]
\footnotesize
\item Confidence intervals and $p$ values are computed by 10,000 paired episode-level bootstrap resamples.
\end{tablenotes}
\end{threeparttable}
\end{table*}

%% file: tables/table2_sample_mechanisms.tex
\begin{table*}[t]
\centering
\begin{threeparttable}
\caption{Sample-level transition statistics for semantic saturation and emergence.}
\label{tab:sample_mechanisms}
\footnotesize
\setlength{\tabcolsep}{2.7pt}
\renewcommand{\arraystretch}{1.08}
\begin{tabular}{llrrrrrl}
\toprule
\multirow{2}{*}{Dataset} & \multirow{2}{*}{Shot} & \multicolumn{2}{c}{Detailed-ZS transitions} & \multicolumn{1}{c}{Cross-view overlap} & \multicolumn{2}{c}{Post-LoRA exclusivity} & \multirow{2}{*}{Regime} \\
\cmidrule(lr){3-4}\cmidrule(lr){5-5}\cmidrule(lr){6-7}
 & & Retention [95\% CI] & Error correction [95\% CI] & Base-LoRA coverage [95\% CI] & D-only & B-only & \\
\midrule
\multirow{2}{*}{EuroSAT} & 1 & 93.9 [93.5, 94.3] & 68.2 [67.2, 69.3] & 84.4 [83.2, 85.6] & 5.8 & 3.6 & \textsc{Sat.} \\
 & 5 & 97.2 [96.9, 97.5] & 87.4 [86.6, 88.2] & 95.2 [94.4, 95.9] & 2.4 & 1.6 & \textsc{Sat.} \\
\addlinespace[2pt]
\multirow{2}{*}{CropDisease} & 1 & 95.4 [95.0, 95.8] & 69.2 [68.0, 70.4] & 86.7 [85.8, 87.7] & 7.3 & 4.3 & \textsc{Sat.} \\
 & 5 & 98.6 [98.3, 98.9] & 90.3 [89.4, 91.2] & 97.3 [96.7, 97.8] & 2.3 & 1.6 & \textsc{Sat.} \\
\addlinespace[2pt]
\multirow{2}{*}{ISIC} & 1 & 53.9 [52.7, 55.1] & 34.0 [33.3, 34.6] & 33.5 [32.1, 35.0] & 11.4 & 7.6 & \textsc{Emer.} \\
 & 5 & 58.3 [57.0, 59.8] & 52.0 [51.0, 52.9] & 48.7 [46.5, 51.0] & 10.9 & 7.8 & \textsc{Emer.} \\
\addlinespace[2pt]
\multirow{2}{*}{ChestX} & 1 & 28.9 [27.7, 30.1] & 20.0 [19.6, 20.5] & 23.0 [21.7, 24.3] & 11.2 & 10.8 & \textsc{Emer.} \\
 & 5 & 33.5 [32.0, 34.8] & 22.2 [21.7, 22.8] & 20.6 [19.1, 22.1] & 13.8 & 12.2 & \textsc{Emer.} \\
\bottomrule
\end{tabular}
\begin{tablenotes}[flushleft]
\footnotesize
\item Retention is the fraction of Detailed-ZS correct samples retained by Detailed-LoRA. Error correction is the fraction of Detailed-ZS errors corrected by Detailed-LoRA. Base-LoRA coverage is measured on samples correct only under Detailed-ZS. D-only and B-only are percentages of all query samples correct exclusively under Detailed-LoRA and Base-LoRA, respectively; their difference equals the adapted semantic utility $\deltalora$.
\end{tablenotes}
\end{threeparttable}
\end{table*}

%% file: tables/table4_shuffle_control.tex
\begin{table*}[t]
\centering
\begin{threeparttable}
\caption{Effect of correct class--text alignment relative to a fixed-point-free shuffled Detailed codebook.}
\label{tab:shuffle_control}
\small
\setlength{\tabcolsep}{4.2pt}
\renewcommand{\arraystretch}{1.08}
\begin{tabular}{llrrrr}
\toprule
\multirow{2}{*}{Dataset} & \multirow{2}{*}{Shot} & \multicolumn{2}{c}{Zero-shot} & \multicolumn{2}{c}{After visual LoRA} \\
\cmidrule(lr){3-4}\cmidrule(lr){5-6}
 & & Aligned $-$ shuffled [95\% CI] & $p$ & Aligned $-$ shuffled [95\% CI] & $p$ \\
\midrule
\multirow{2}{*}{EuroSAT} & 1 & +53.64 [+50.88, +56.41] & $<.0001$ & +36.92 [+34.08, +39.69] & $<.0001$ \\
 & 5 & +53.33 [+50.68, +56.00] & $<.0001$ & +15.47 [+13.92, +17.07] & $<.0001$ \\
\addlinespace[2pt]
\multirow{2}{*}{CropDisease} & 1 & +53.20 [+49.72, +56.60] & $<.0001$ & +28.25 [+25.88, +30.60] & $<.0001$ \\
 & 5 & +53.48 [+49.91, +56.95] & $<.0001$ & +4.64 [+3.72, +5.61] & $<.0001$ \\
\addlinespace[2pt]
\multirow{2}{*}{ISIC} & 1 & +8.85 [+7.63, +10.05] & $<.0001$ & +10.88 [+9.16, +12.56] & $<.0001$ \\
 & 5 & +8.71 [+7.43, +10.03] & $<.0001$ & +8.97 [+7.36, +10.56] & $<.0001$ \\
\addlinespace[2pt]
\multirow{2}{*}{ChestX} & 1 & +1.61 [+0.45, +2.76] & 0.007 & +0.63 [-0.53, +1.75] & 0.291 \\
 & 5 & +0.99 [-0.01, +2.01] & 0.053 & +1.15 [-0.11, +2.39] & 0.072 \\
\bottomrule
\end{tabular}
\begin{tablenotes}[flushleft]
\footnotesize
\item The shuffled condition preserves the same Detailed text-embedding multiset but permutes its class assignment within each episode. Positive values favor correct semantic alignment. Confidence intervals and two-sided $p$-values use 10,000 paired episode-level bootstrap resamples.
\end{tablenotes}
\end{threeparttable}
\end{table*}

%% file: tables/table5_backbone_generalization.tex
\begin{table*}[t]
\centering
\begin{threeparttable}
\caption{Backbone generalization of adaptation-conditional semantic utility.}
\label{tab:backbone_generalization}
\footnotesize
\setlength{\tabcolsep}{2.8pt}
\renewcommand{\arraystretch}{1.08}
\begin{tabular}{lllrrrrrrl}
\toprule
\multirow{2}{*}{Dataset} & \multirow{2}{*}{Backbone} & \multirow{2}{*}{Shot} & \multicolumn{2}{c}{Zero-shot} & \multicolumn{2}{c}{After visual LoRA} & \multicolumn{2}{c}{Semantic utility (pp)} & \multirow{2}{*}{Regime} \\
\cmidrule(lr){4-5}\cmidrule(lr){6-7}\cmidrule(lr){8-9}
 & & & Base & Detailed & Base & Detailed & $\Delta_0$ & $\Delta_L$ & \\
\midrule
\multirow{4}{*}{EuroSAT} & ViT-B/16 & 1 & 54.85 & \textbf{62.98} & 82.19 & \textbf{84.40} & +8.13 & +2.21 & \textsc{Sat.} \\
 & ViT-B/16 & 5 & 54.55 & \textbf{62.86} & 92.74 & \textbf{93.55} & +8.31 & +0.82 & \textsc{Sat.} \\
 & ViT-B/32 & 1 & 48.12 & \textbf{56.80} & 79.49 & \textbf{82.84} & +8.68 & +3.35 & \textsc{Sat.} \\
 & ViT-B/32 & 5 & 48.32 & \textbf{57.05} & 92.32 & \textbf{93.37} & +8.73 & +1.05 & \textsc{Sat.} \\
\addlinespace[2pt]
\multirow{4}{*}{CropDisease} & ViT-B/16 & 1 & 40.90 & \textbf{61.91} & 82.48 & \textbf{85.44} & +21.00 & +2.96 & \textsc{Sat.} \\
 & ViT-B/16 & 5 & 40.80 & \textbf{62.34} & 94.80 & \textbf{95.49} & +21.54 & +0.69 & \textsc{Sat.} \\
 & ViT-B/32 & 1 & 33.92 & \textbf{49.49} & 77.92 & \textbf{82.00} & +15.57 & +4.08 & \textsc{Sat.} \\
 & ViT-B/32 & 5 & 33.92 & \textbf{49.33} & 93.16 & \textbf{94.97} & +15.41 & +1.81 & \textsc{Sat.} \\
\addlinespace[2pt]
\multirow{4}{*}{ISIC} & ViT-B/16 & 1 & \textbf{27.24} & 26.50 & 35.43 & \textbf{39.27} & -0.74 & +3.84 & \textsc{Emer.} \\
 & ViT-B/16 & 5 & \textbf{27.08} & 26.57 & 50.55 & \textbf{53.67} & -0.51 & +3.12 & \textsc{Emer.} \\
 & ViT-B/32 & 1 & 27.12 & \textbf{27.48} & 36.40 & \textbf{39.28} & +0.36 & +2.88 & \textsc{Ampl.} \\
 & ViT-B/32 & 5 & \textbf{27.15} & 26.85 & 50.03 & \textbf{52.19} & -0.29 & +2.16 & \textsc{Emer.} \\
\addlinespace[2pt]
\multirow{4}{*}{ChestX} & ViT-B/16 & 1 & \textbf{21.93} & 21.18 & 21.50 & \textbf{21.90} & -0.75 & +0.40 & \textsc{Emer.} \\
 & ViT-B/16 & 5 & \textbf{21.84} & 21.17 & 23.06 & \textbf{24.60} & -0.67 & +1.54 & \textsc{Emer.} \\
 & ViT-B/32 & 1 & \textbf{20.08} & 19.37 & 21.23 & \textbf{22.56} & -0.71 & +1.33 & \textsc{Emer.} \\
 & ViT-B/32 & 5 & \textbf{20.04} & 18.93 & 21.51 & \textbf{23.44} & -1.11 & +1.93 & \textsc{Emer.} \\
\bottomrule
\end{tabular}
\begin{tablenotes}[flushleft]
\footnotesize
\item Bold marks the better text view for the same backbone and adaptation state. Regimes follow the sign definitions in Section~\ref{sec:problem}; \textsc{Ampl.} denotes positive utility that increases after adaptation.
\end{tablenotes}
\end{threeparttable}
\end{table*}

%% file: tables/table6_multiseed_summary.tex
\begin{table*}[t]
\centering
\begin{threeparttable}
\caption{Three-seed robustness of semantic utility on ViT-B/16.}
\label{tab:multiseed_summary}
\small
\setlength{\tabcolsep}{5.0pt}
\renewcommand{\arraystretch}{1.08}
\begin{tabular}{llrrrrl}
\toprule
Dataset & Shot & $\Delta_0$ & $\Delta_L$ & $\Delta_{\mathrm{shift}}$ & Regime stable & Majority regime \\
\midrule
\multirow{2}{*}{EuroSAT} & 1 & +8.02 $\pm$ 0.18 & +2.15 $\pm$ 0.51 & -5.87 $\pm$ 0.35 & \checkmark & \textsc{Sat.} \\
 & 5 & +8.57 $\pm$ 0.30 & +0.94 $\pm$ 0.17 & -7.63 $\pm$ 0.13 & \checkmark & \textsc{Sat.} \\
\addlinespace[2pt]
\multirow{2}{*}{CropDisease} & 1 & +21.66 $\pm$ 0.58 & +3.03 $\pm$ 0.25 & -18.63 $\pm$ 0.52 & \checkmark & \textsc{Sat.} \\
 & 5 & +22.65 $\pm$ 0.97 & +0.80 $\pm$ 0.14 & -21.84 $\pm$ 0.86 & \checkmark & \textsc{Sat.} \\
\addlinespace[2pt]
\multirow{2}{*}{ISIC} & 1 & -0.88 $\pm$ 0.35 & +3.62 $\pm$ 0.74 & +4.50 $\pm$ 0.40 & \checkmark & \textsc{Emer.} \\
 & 5 & -0.75 $\pm$ 0.22 & +4.05 $\pm$ 0.87 & +4.80 $\pm$ 1.09 & \checkmark & \textsc{Emer.} \\
\addlinespace[2pt]
\multirow{2}{*}{ChestX} & 1 & -0.62 $\pm$ 0.15 & +0.26 $\pm$ 0.29 & +0.87 $\pm$ 0.28 & -- & \textsc{Emer.} \\
 & 5 & -0.46 $\pm$ 0.25 & +1.67 $\pm$ 0.47 & +2.12 $\pm$ 0.31 & \checkmark & \textsc{Emer.} \\
\bottomrule
\end{tabular}
\begin{tablenotes}[flushleft]
\footnotesize
\item Entries are mean $\pm$ standard deviation across three episodic seeds. A checkmark indicates that the exact sign-based regime is identical for all seeds. ChestX 1-shot is a weak boundary case whose adapted utility is close to zero.
\end{tablenotes}
\end{threeparttable}
\end{table*}

%% file: tables/table3_prompt_transfer.tex
\begin{table}[t]
\centering
\caption{Prompt-conditioned co-adaptation measured by cross-prompt evaluation.}
\label{tab:prompt_transfer}
\scriptsize
\setlength{\tabcolsep}{2.0pt}
\renewcommand{\arraystretch}{1.06}
\resizebox{\columnwidth}{!}{%
\begin{tabular}{llrrrrrr}
\toprule
\multirow{2}{*}{Dataset} & \multirow{2}{*}{Shot} & \multicolumn{2}{c}{Train Base} & \multicolumn{2}{c}{Train Detailed} & \multicolumn{2}{c}{Gap (pp)} \\
\cmidrule(lr){3-4}\cmidrule(lr){5-6}\cmidrule(lr){7-8}
 & & $B\!\to\!B$ & $B\!\to\!D$ & $D\!\to\!B$ & $D\!\to\!D$ & $G_B$ & $G_D$ \\
\midrule
\multirow{2}{*}{EuroSAT} & 1 & 82.19 & 79.70 & 77.46 & 84.40 & +2.49 & +6.94 \\
 & 5 & 92.74 & 89.56 & 89.61 & 93.55 & +3.18 & +3.94 \\
\addlinespace[1pt]
\multirow{2}{*}{CropDisease} & 1 & 82.48 & 68.32 & 63.22 & 85.44 & +14.16 & +22.22 \\
 & 5 & 94.80 & 80.13 & 79.40 & 95.49 & +14.67 & +16.09 \\
\addlinespace[1pt]
\multirow{2}{*}{ISIC} & 1 & 35.43 & 38.06 & 33.61 & 39.27 & -2.63 & +5.66 \\
 & 5 & 50.55 & 48.96 & 47.59 & 53.67 & +1.59 & +6.08 \\
\addlinespace[1pt]
\multirow{2}{*}{ChestX} & 1 & 21.50 & 20.26 & 21.59 & 21.90 & +1.24 & +0.31 \\
 & 5 & 23.06 & 20.17 & 23.71 & 24.60 & +2.89 & +0.89 \\
\bottomrule
\end{tabular}}
\vspace{1pt}
\parbox{\columnwidth}{\footnotesize $B$ and $D$ denote Base and Detailed prompts. $G_B$ and $G_D$ are signed matched-minus-cross-prompt accuracy differences; positive values favor the training prompt.}
\end{table}

%% file: tables/table7_contextual_comparison.tex
% Intentionally non-floating: the contextual comparison must remain next to its
% qualification text instead of being deferred into the bibliography.
\refstepcounter{table}\label{tab:contextual_comparison}
\begin{center}
{\footnotesize TABLE~\thetable\\[-1pt]
Contextual 5-way 1-shot comparison (\%).}\\[4pt]
\scriptsize
\setlength{\tabcolsep}{2.0pt}
\renewcommand{\arraystretch}{1.05}
\resizebox{\columnwidth}{!}{%
\begin{tabular}{lrrrrr}
\toprule
Method & ChestX & ISIC & EuroSAT & CropDisease & Avg. \\
\midrule
IM-DCL~\cite{xu2024imdcl} & 23.98 & 38.13 & 77.14 & 84.37 & 55.91 \\
StepSPT~\cite{xu2025stepspt} & 22.84 & 32.97 & 70.01 & 84.84 & 52.68 \\
CLIP-LoRA (reported)~\cite{zanella2024clip_lora,yi2026atha} & 21.73 & 35.23 & 81.41 & 85.32 & 55.92 \\
Semantic Probe (LoRA)~\cite{zhao2026reviving} & 23.65 & 38.77 & 82.94 & 85.11 & 57.62 \\
ATHA~\cite{yi2026atha} & \textbf{24.00} & 38.86 & 82.56 & \textbf{87.99} & \textbf{58.35} \\
\textbf{Detailed-LoRA (ours)} & 21.90 & \textbf{39.27} & \textbf{84.40} & 85.44 & 57.75 \\
\bottomrule
\end{tabular}}
\end{center}
\vspace{2pt}

\refstepcounter{table}\label{tab:contextual_comparison_5shot}
\begin{center}
{\footnotesize TABLE~\thetable\\[-1pt]
Contextual 5-way 5-shot comparison (\%).}\\[4pt]
\scriptsize
\setlength{\tabcolsep}{2.0pt}
\renewcommand{\arraystretch}{1.05}
\resizebox{\columnwidth}{!}{%
\begin{tabular}{lrrrrr}
\toprule
Method & ChestX & ISIC & EuroSAT & CropDisease & Avg. \\
\midrule
IM-DCL~\cite{xu2024imdcl} & \textbf{28.93} & 52.74 & 89.47 & 95.73 & 66.72 \\
StepSPT~\cite{xu2025stepspt} & 26.36 & 52.12 & 89.40 & 96.01 & 65.97 \\
CLIP-LoRA (reported)~\cite{zanella2024clip_lora,yi2026atha} & 24.13 & 51.10 & 92.52 & 96.21 & 65.99 \\
Semantic Probe (LoRA)~\cite{zhao2026reviving} & 25.79 & 55.95 & 93.43 & 96.88 & 68.01 \\
ATHA~\cite{yi2026atha} & 26.67 & \textbf{56.42} & 93.41 & \textbf{97.62} & \textbf{68.53} \\
\textbf{Detailed-LoRA (ours)} & 24.60 & 53.67 & \textbf{93.55} & 95.49 & 66.83 \\
\bottomrule
\end{tabular}}
\end{center}
\vspace{2pt}

\noindent\footnotesize\emph{Comparison note:} values are transcribed from the cited papers. They are contextual rather than harmonized because backbones, episode counts, optimization budgets, augmentation, and language resources differ. Bold marks the best displayed value within a column and is not a significance claim.\normalsize